\documentclass[lettersize,journal]{IEEEtran}
\usepackage{amsmath,amsfonts}
\usepackage{algorithmic}
\usepackage{algorithm}
\usepackage{xcolor}
\usepackage{array}
\usepackage[caption=false,font=normalsize,labelfont=sf,textfont=sf]{subfig}
\usepackage{textcomp}
\usepackage{stfloats}
\usepackage{url}
\usepackage{verbatim}
\usepackage{graphicx}
\usepackage{cite}
\newcommand{\mD}{\mathcal{D}}
\newcommand{\mR}{${\rm I\!R}$}
\usepackage{xcolor}
\begin{document}

\title{Mapping Temporary Slums from Satellite Imagery using a Semi-Supervised Approach}
\author{M. Fasi ur Rehman, Izza Ali, Waqas Sultani, Mohsen Ali
\thanks{M. Fasi ur Rehman and Izza Ali are with the School of Humanities and Social Sciences and Waqas Sultani and Mohsen Ali are with the department of computer science, Information Technology University, Lahore. Email: waqas.sultani@itu.edu.pk, mohsen.ali@itu.edu.pk  }}

\markboth{Journal of \LaTeX\ Class Files,~Vol.~14, No.~8, August~2021}%
{Shell \MakeLowercase{\textit{et al.}}: A Sample Article Using IEEEtran.cls for IEEE Journals}


\maketitle

\begin{abstract}
 
One billion people worldwide are estimated to be living in slums, and documenting and analyzing these regions is a challenging task.  
As compared to regular slums; the small, scattered and temporary nature of temporary slums makes data collection and labeling tedious and time-consuming. 
To tackle this challenging problem of temporary slums detection, we present a semi-supervised deep learning segmentation-based approach; with the strategy to detect initial seed images in the zero-labeled data settings.
A small set of seed samples (32 in our case) are automatically discovered by analyzing the temporal changes, which are manually labeled to train a segmentation and representation learning module.
The \textit{segmentation module} gathers high dimensional image representations, and the representation \textit{learning module} transforms image representations into embedding vectors. After that,  a scoring module uses the embedding vectors to sample images from a large pool of unlabeled images and generates pseudo-labels for the sampled images. These sampled images with their pseudo-labels are added to the training set to update the segmentation and representation learning modules iteratively. 
To analyze the effectiveness of our technique, we construct a large geographically marked dataset of temporary slums. 
This dataset constitutes more than 200 potential temporary slum locations (2.28 square kilometers) found by sieving sixty-eight thousand images from 12 metropolitan cities of Pakistan covering 8000 square kilometers. Furthermore, our proposed method outperforms several competitive semi-supervised semantic segmentation baselines on a similar setting.  The code and the dataset will be made publicly available.
\end{abstract}


\section{Introduction}
\IEEEPARstart{S}{lums} are home to about a billion humans worldwide \cite{slumAlmanac}. The lack of opportunities in the rural areas pushes people towards urban areas; thus, two-thirds of the global population is expected to be living in urban areas by 2050 \cite{un2015}. 
 
\begin{figure*}[!t]
\centering
  \includegraphics[width=.7\textwidth]{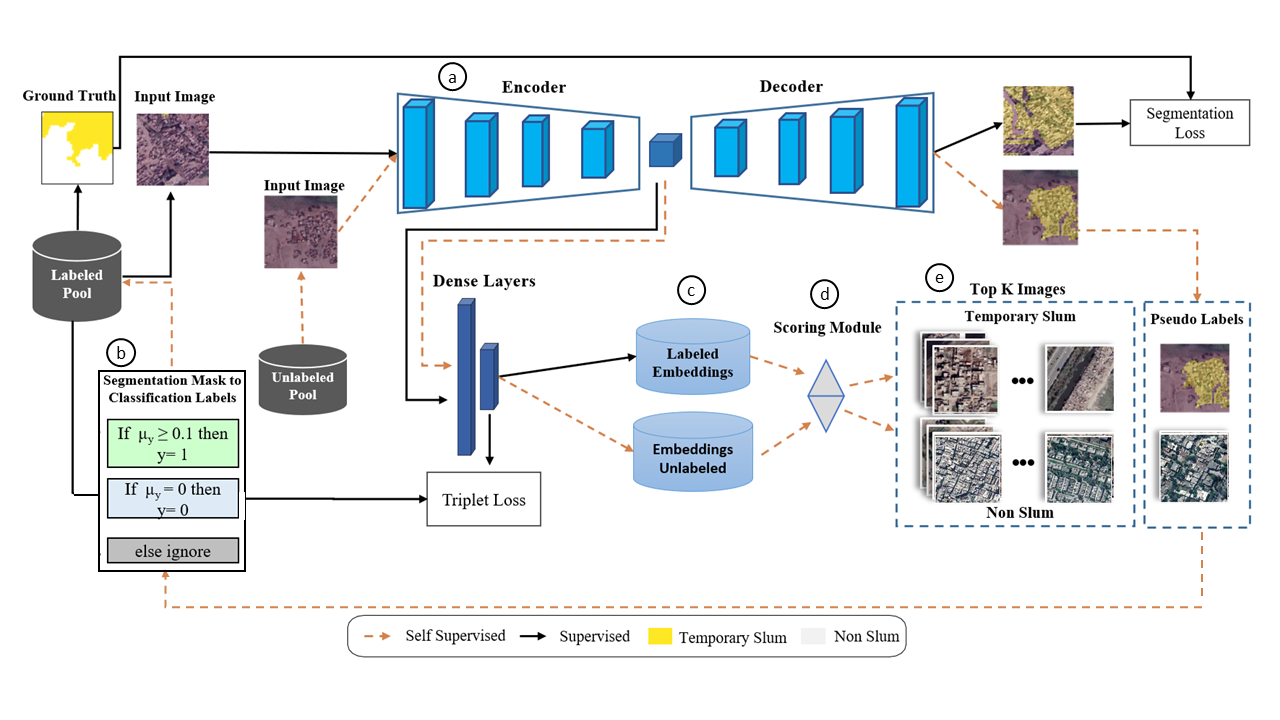}
   \caption{ {(a) We have used U-Net architecture as a segmentation model. Two one-dimensional fully connected layers of 256 and 64 units are appended at the end of the centermost layer of U-Net. The arrows from the centermost layer of U-Net to the dense layers indicate the flow of feature maps. (b) Converting pixel-level labels to image-level labels: y=1 means the image is labeled as a slum. In contrast, y=0 means the image is labeled as non-slum, and $\mu_{y}$ is the mean value of the segmentation mask or the ratio of area covered by pixels labeled as a slum. (c) represent 64-dimensional embedding for labeled and unlabeled data. (d) The scoring module is based on Cosine similarity between labeled and unlabeled embeddings. (e) The scoring module returns two lists of the sorted list, one for each class i.e, slum and non-slum. Top K images are selected from each list.} \vspace{-0.2cm}}\label{figure1}
\end{figure*}

\begin{figure}[!t]
\centering
  \includegraphics[width=.45\textwidth]{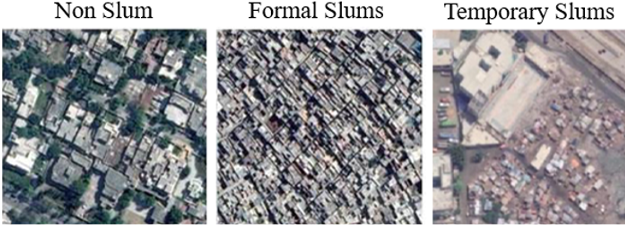}
  
  \caption{Examples of samples from Lahore, Pakistan.}\label{figure2}
\end{figure}

.
Slums can be detected manually \cite{slumanual1} and automatically using object-based image analysis (OBIA) \cite{slumauto1, slumauto2} and by employing a grey-level-co-occurrence-matrix and support vector machines \cite{slumGLCM1, slumSVM1}. Recently deep learning-based slums detection approaches have been proposed, including  \cite{slumDeep, wurm2019semantic}. Wurm et al., \cite{simpleslum} used fully convolutional network 
architecture to segment out slum areas from non-slum. Similarly, authors in  \cite{wurm2019semantic}, and \cite{imbSlums} showed that the slum mapping-related information could be transferred between models trained on different image resolutions, thus improving the segmentation quality across the board. 

 {
UN-Habitat  \cite{desa2016transforming} defines a slum by the lack of one or more of the following: durable housing, sufficient living space, easy access to safe water, access to adequate sanitation, and security of tenure. Normal/formal slum areas, although unplanned, due to their long-time existence, take the shape of semi-formal structure with a population stretching to thousands. Many of these have some partial local economic activity and formal sources of basic amenities, electricity, and water \cite{mustafainformal}. The \textit{temporary slums}, as the name suggests, are temporary both in the context of existence (time) and the building structures present there. Due to their temporary nature, the area lacks much of the formal structure, access to amenities and mainly consists of buildings that could be moved easily (make-shift houses, tents). Therefore, temporary slum dwellers are even more marginalized and economically and socially more divested. The unplanned nature of the development in these dwellings makes human-centered information gathering an expansive exercise. The distinction between temporary (informal) and formal slums is often overlooked. We could not find any reliable source of geographical locations of temporary slums and the number of people living in temporary slums around the world. Through this work, we intend to provide geographical locations of temporary slums in Pakistan and a framework that can be used in other parts of the world so we can be made aware of the scale of temporary slums.
}

Our work is focused specifically on identifying temporary slums from satellite imagery (examples of temporary slums are shown in Fig. \ref{figure2}). One of the challenging aspects of detecting these structures is the lack of data in terms of visual datasets and location information. Temporary slums are very small and scattered around the cities; deciding which images to label or, more simply put, "where to look for temporary slums" is a \textcolor{black}{vital issue}. 
 
To overcome the limitation of the availability of the geo-located datasets, we have designed a technique to collect \textit{seed images} exploiting the temporary nature of these slums and design a \textit{semi-supervised learning} approach to segment out the slums in the region. Our methodology allows it to be applicable to any region since it learns from the structure in the local images. Specifically, our work addresses the challenge of finding temporary slums in an automated way from visible spectrum (RGB) satellite imagery by learning from a minimal dataset and then iteratively extending the dataset. 
In summary, our work has the following contributions: (1) A  semi-supervised semantic-segmentation method is presented that learns to identify and segment out the temporary slum from very few positive seed samples, (2) We devise a strategy to start from a zero-dataset scenario. The make-shift nature of the temporary slums is exploited to discover the small set of seed images that might contain the temporary slum areas, (3) We put forward a new large and diverse satellite imagery-based dataset of temporary slums location. The total footprint of the collected dataset is 2.28 square kilometers and covers all 12 metropolitan cities of Pakistan, (4) The proposed approach outperforms several competitive baselines. Specifically, it outperforms supervised (trained on seed images) baseline by 1.8 times and semi-supervised baselines by 1.9 times.

\section{Methodology}

 
 {
Our proposed approach starts with collecting seed satellite images representing temporary slums for the semi-supervised learning task. To overcome the limitation of information regarding where temporary slums are located, we design an automatic strategy to collect seed images using temporal changes in the status of buildings. The semi-supervised framework of extending the seed dataset (labeled pool) is illustrated in  (Fig. \ref{figure1}). Using the seed dataset, we train a segmentation model and an encoding model. The encoding model is then used to extract encoding from both labeled and unlabeled images. Encoding of slum labeled images are compared will all of the embeddings of the unlabeled pool and are sorted based on cosine similarity, and the same process is repeated for non-slum labeled images. As a result, for each class (slum and non-slum), we have a sorted list of images that are similar to the relative class. We select the top few (top $K$) images from each list and generate segmentation masks.  If the segmented region is not of significant size, it is removed from the top list. The remaining images in the top $K$ list along with their segmentation masks are made a part of the seed dataset.}

\subsection{Preliminaries}
We start with two set of images $\mD_{pool}$ and $\mD_{lab}$. 
$\mD_{pool}$ consist of unlabeled images $X_{pool}$, whereas $\mD_{lab}$  consist of labeled images $X_{lab}$ with segmentation labels $Y_{lab}$. 
Let $X_{pseudo}$ be the images taken from the pool $\mD_{pool}$ and added to the $\mD_{lab}$ over multiple iterations of the self-supervised learning algorithm. 
Each image in \(X_{pool}\), \(X_{lab}\) and \(X_{pseudo}\) are such that \(x \in \mR^{W\times H}\) where $W\times H$ are the spatial dimensions of the images. 
Our segmentation model \(f_{s}\) 
outputs, \(p = f_{s}(x)\), probability of each pixel belonging to the temporary slum. 
We design an embedding model \(f_{emb}\) which consists of the encoder part of \(f_{s}\) extended with fully connected layers. 
The output of this \(f_{emb}\) model is $\mathcal{E}(x) \in \mR^{64\times 1}$. 
Unlike many previous works in semi-supervised learning, we start with empty $\mD_{lab}$ and, using our temporal consistency strategy we select a few images which are hand-tagged to construct initial $\mD_{lab}$. 
Over time, both the \(f_{s}\) and \(f_{emb}\)  are trained jointly over the labeled and pseudo-labeled data.

\subsection{Discovering Seed Images}
\label{sec:discoverSeedImages}

\begin{figure}[t]
\centering
  \includegraphics[width=0.5\textwidth]{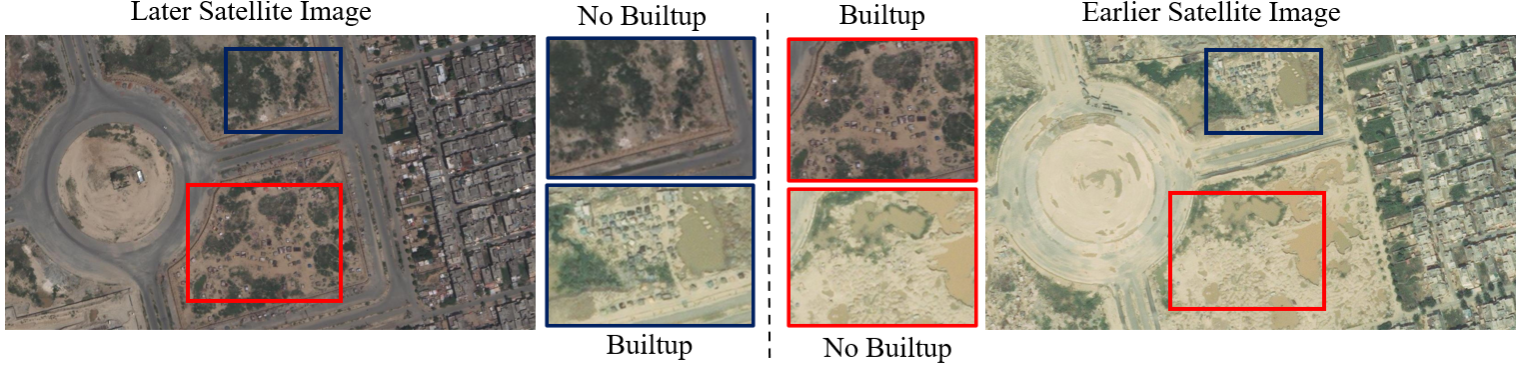}
  \vspace{-0.5cm }
  \caption{An example of an area where the built-up area changed over time. The relative difference between the two Satellite Images taken at two different Timestamps is visible in the zoomed-in views on the right. The area zoomed in blue had no built-up, while some dwellings appear in the later view and opposite for the area zoomed in red.}\label{figure3}
\end{figure}

 We design an automatic strategy to overcome the limitation of the non-existence of geo-location information about the temporary slums. Specifically, we use the non-permanent presence of temporary slums as a means of identifying them with the help of a built-up region segmentation model. A U-Net-based segmentation model is trained over the Village Finder \cite{VF} dataset for the build-up segmentation. We use the change in segmentation results (intersection over union) to classify images where the change in build-up regions occurred; either new settlements were created, or settlements were removed. Any regions with non-nucleated settlement detection or having detected settlements lower than a certain threshold were removed. Working conservatively, only a few images with small IoU across the time are selected as possible temporary slums and pixel-level tagging is performed (Fig. \ref{figure3}). 
 
\textcolor{black}{Change detection might fail to capture all the temporary slums, since they might not change in the time window we choose for this purpose, and might capture images that are not temporary slums.}
 Therefore, this method results in only a few seed images 32 in our case  {which were curated and annotated manually, this process was done for just one city (Lahore)}.
 
 \subsection{Dataset}

The satellite imagery is collected in patches of 256 $\times$ 256 pixels using a sliding window with an overlap of 56 pixels. The unlabeled pool consists of 68,051 images, while the labeled pool contains 32  temporary slum and  4831  non-temporary slum images, which are then split into training (20 slums and 3938 non-slum) and validation (12 slums and 861 slums).  {The seed-image dataset is highly imbalanced (against temporary slums) and extremely small to train a robust semantic-segmentation network. To cater to the data imbalance, we use data augmentations where slum images are augmented more times than non-slum images. To enrich our negative class, we make sure that we add examples from `formal slums' or permanent slums as non-temporary slum examples (Fig. \ref{figure2})}. The source of satellite imagery is Google Earth at zoom level 19, which translates to approximately 1m per pixel resolution. Rawalpindi, Islamabad, Multan, Hyderabad, Faisalabad, Gujranwala, Sialkot, Peshawar, Quetta, Sargodha are part of the unlabeled set (whole city), Lahore is part of the training set (train and Val), whereas Karachi is a part of the test set. \vspace{-.1in}

\subsection{Temporary Slum segmentation Module}
To achieve the pixel-wise location of temporary slums regions, we employ a semantic segmentation module. Our segmentation model \(f_{s}\) is based on U-Net \cite{unet} architecture with the ResNet-32 \cite{resnet} backbone pre-trained on ImageNet \cite{imagenet}. It is trained on \(D_{lab}\) which is our initial labeled dataset in a fully supervised fashion.  
 The U-Net is trained through binary cross-entropy loss.%

\subsection{Embedding Learning Module}

To learn a more robust and discriminative embedding, we have employed deep metric learning. The goal is to push the model into squeezing discriminative semantic information in a one-dimensional vector, commonly referred to as embedding vector, by penalizing the model if the representation of the instances of two different classes is similar. To generate embeddings,  we add three fully connected layers of size $512$, $256$, and $64$ neurons, respectively, over the features extracted from the encoder. 
ReLu activation is used after the first two layers, whereas the third one (output layer) is without any activation layer. 
  
The output of this module is a $\mathcal{E}(I) \in \mR^{64\times 1}$ dimensional embedding-vector of the image $I$. 
We use triplet loss as an objective function.
\begin{equation}
  \mathcal{L}_{emb} = max(\lVert (\mathcal{E}(I_{a}), \mathcal{E}(I_{p}) \lVert ) - \lVert  (\mathcal{E}(I_{a}), \mathcal{E}(I_{n}))\lVert  + m ,0),
\end{equation} 
where  $\mathcal{E}(I_a)$, $\mathcal{E}(I_{p})$,
$\mathcal{E}(I_{n})$ represents embedding vector for the anchor, positive and negative images, and  $m$ is a margin. Positive and anchor images belong to the same class (temporary slum), whereas the negative images are from the non-temporary slum areas.
Minimizing the triplet loss forces the embedding of images from the temporary slum areas to be similar while maximizing the distance between embedding from temporary slum and non-temporary slum areas. The triplets are created on a fly from the batch. 

\subsection{Discovering Temporary Slum from Unlabelled Pool}

We start with  $\mD_{pool}$ which consist of unlabeled images $X_{pool}$ and $\mD_{lab}$  which consist of labeled images $X_{lab}$ with segmentation labels $Y_{lab}$. 
At each iteration of our algorithm, selected samples from $\mD_{pool}$ are moved to the $\mD_{lab}$ along with their pseudo-labels. 
Let $X_{pseudo}$ be such images. 
Each image in \(X_{pool}\),\(X_{lab}\) and \(X_{pseudo}\) are such that \(x_{s} \in \mR^{W\times H}\) where $W$ and $H$ are the spatial dimensions of the images.

\subsubsection{Scoring Module}
\begin{figure}
\centering
  \includegraphics[width=.45\textwidth]{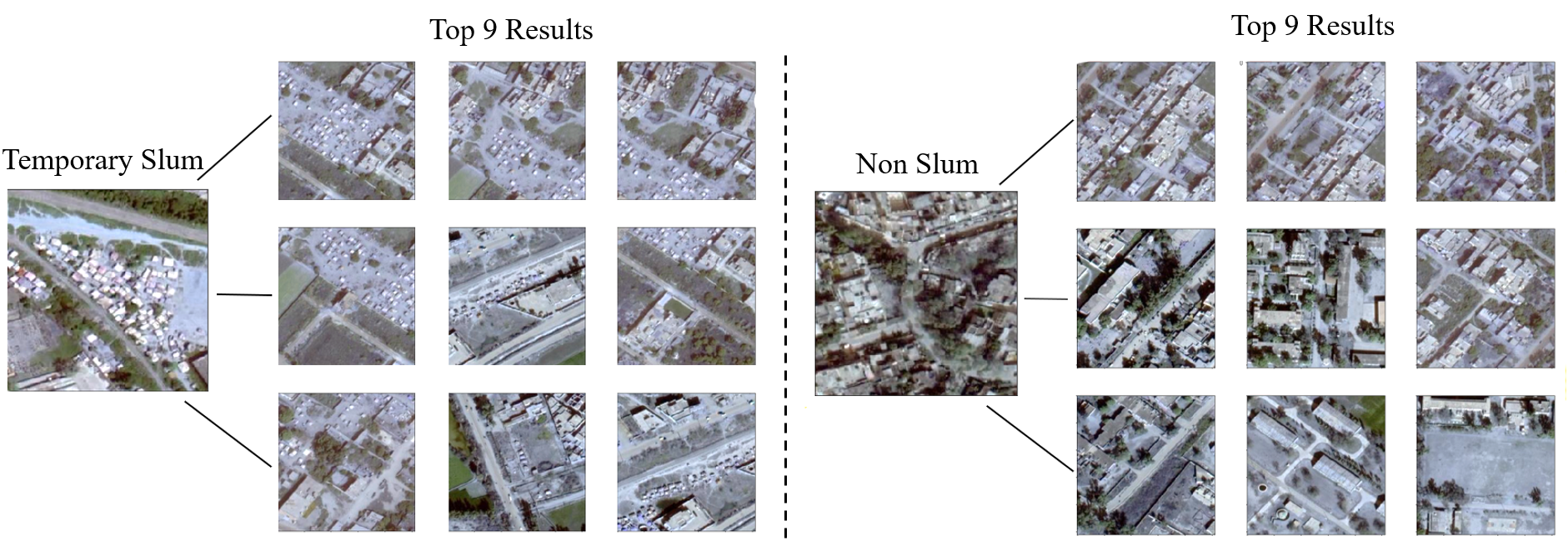}
  \caption{Image retrieval using cosine similarity. Image retrieval is done by comparing the labeled dataset's embedding vectors with the unlabeled dataset's embedding vectors. Top 9 Query results for temporary slum examples are shown on the left, and for non-slum, examples are shown on the right.}\label{figure4}
\end{figure}

The trained semantic segmentation $f_s$ network with embedding head is used to generate the image embedding for all the images in the unlabeled pool as well as the labeled pool, $X_{pool} \cup X_{lab}$. 
We compute the cosine similarity between embeddings across the pool to identify unlabeled pool images similar to images in the labeled pool.  Specifically, embedding of each unlabeled image $x_p^{i} \in \mathcal{X}_{pool}$  is compared with the embedding of all slum and non-slum images $x_{lab}^{i} \in \mathcal{X}_{lab}$ from the labeled pool.
A few examples are shown in  Fig. \ref{figure4}.
Instead of calculating pixel-level similarity, calculating similarity over the image embedding allows us to compare the encoded information in the latent space. Let $K_p^t$ and  $K_l^t$ be the total number of images in $\mathcal{X}_{pool}$ and $\mathcal{X}_{lab}$ at iteration $t$. 
 Let $d(i_p, i_l)= \mathcal{C}(\mathcal{E}(x_p^i), \mathcal{E}(x_l^i))$ be the cosine similarity between embeddings of  $x_p^{i} \in \mathcal{X}_{pool}$ and $x_{l}^{i} \in \mathcal{X}_{lab}$.
 Let $\mu_{s}^{i}$ be the average similarity of $x_p^{i}$ with all the temporary slum images in the $\mathcal{X}_{lab}$ and $\mu_{n}^{i}$ be the average similarity with all the non-slum images in the $\mathcal{X}_{lab}$.  $\mu_{s}^{i}$ and $\mu_{n}^{i}$ are sorted in descending order and top $K$ images selected to make sets $S_s^t$ and $S_{ns}^t$. Images that get selected in both sets are removed to make exclusive sets and are added back to the unlabeled pool.  

\subsubsection{Pixel level Pseudo Labeling}
 After $S_s^t$ and $S_{ns}^t$ have been selected based on image-level information, the second step of filtering is performed, this time over the pixel-level information. 
We compute the segmentation mask using the model $f_s$ updated in the last iteration for all the images in the selected sets.
Images in  $S_s^t$, images with predicted temporary slum segmentation area less than 5 percent of the total image size are removed from the set. While for images from non-temporary slum set  $S_{ns}^t$ if the slum class prediction covers more than 5 percent area, they are removed from the set. The remaining images from each set are added to the labeled data pool along with their segmentation pseudo labels, which are computed by applying thresholds to prediction probabilities at 0.5.
 
\begin{figure}
\centering
  \includegraphics[width=.45\textwidth]{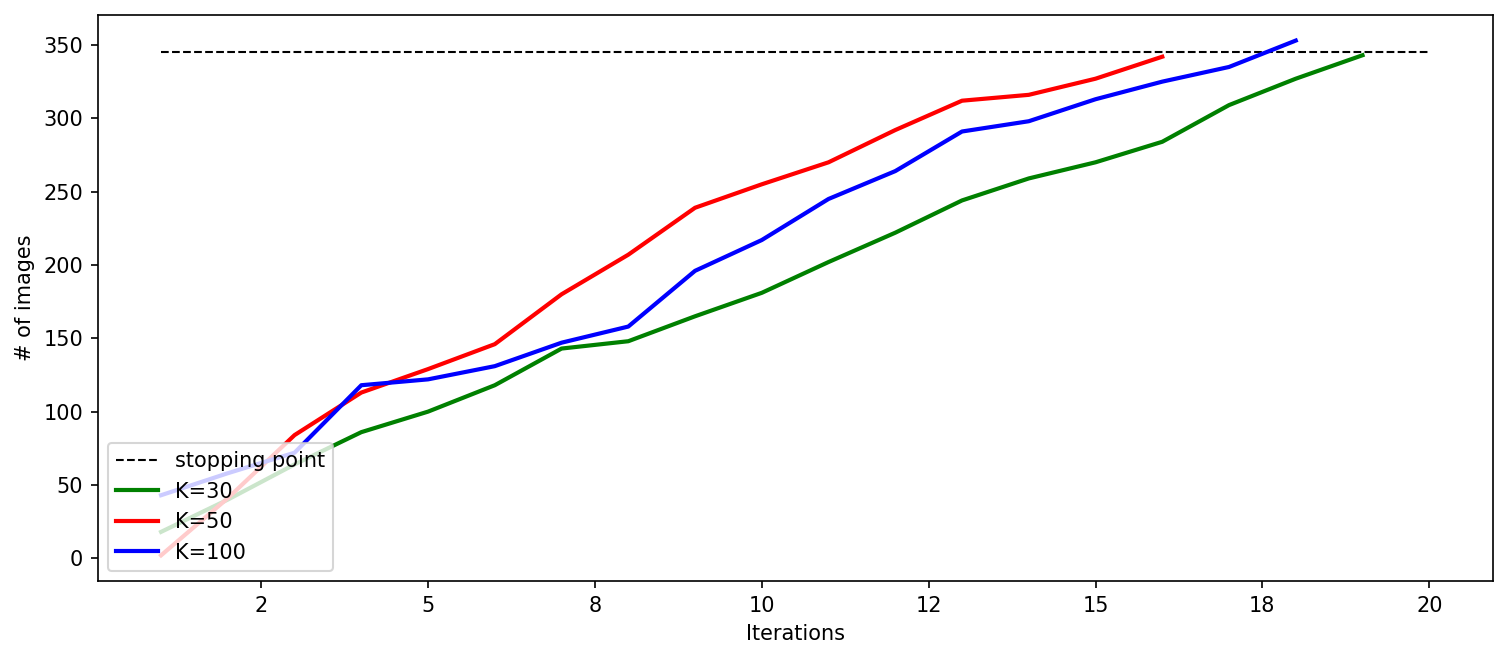}
  \caption{Iterations taken to reach desired set of images by algorithm on different K settings. Stopping point shown as dashed line in black. The solid lines show the cumulative count of images added to $S_s^t$ over the course of the experiment.}\label{stopping}
\end{figure}

\section{Experiments and Results}
\subsection{Experimental Setup}\label{ExperimentalSetup}
Segmentation model \(f_{s}\) and embedding model \(f_{e}\) are trained for at most 50 epochs with batch size 8 and learning rate of \(10^{-3}\) and \(10^{-5}\) respectively.  
Since the proposed method is an iterative process, the stopping criterion is defined as follows; when the ratio of the size of $S_s^t$ to the initial size of the unlabeled pool $\mD_{pool}$ becomes equal to (or greater than) ratio of \textit{slums to non-slums} as present in the initial training set, no further iterations are performed. This criterion is based on the simple intuition that the slum to the non-slum ratio in the initial training set is indicative of the actual presence of slums. 
Our initial training set consists of $3938$ images in which
$32$ ($20$ used for training and $10$ used for validation) are labeled as slums while the rest are labeled non-slum.  In other words, only $0.507\%$ of the training pool is labeled as a slum. In this case, to reach the same ratio, we have to identify at least 345 slum images from the unlabeled pool. Our test set 486 images, of which $196$ are labeled as temporary slums.

\begin{table}[t]
  \caption{ 
Quantitative Comparison of a few semi supervised learning base semantic segmentation methods \label{Tab:Tcr}}
\centering
\begin{tabular}{|l|l|l|l|l|}

\hline
\textit{Method} & \textit{mIoU} & \textit{Precision} & \textit{Recall} & \textit{F1} \\ \hline
Supervised baseline (U-Net) & 0.17 & 0.27 & 0.27 & 0.22  \\ \hline
Cutmix\cite{cutmix} & 0.11 & 0.35 & 0.13 & 0.16\\ \hline
Cutout\cite{cutmix} & 0.14 & 0.39 & 0.18 & 0.19  \\ \hline
ICT\cite{ict}  & 0.18 & 0.38 & 0.24 & 0.25 \\ \hline
VAT\cite{VAT} & 0.18 & 0.40 & 0.23 & 0.24\\ \hline
\textbf{Ours} & 0.33 & 0.49 & 0.49 & 0.43\\ \hline

\end{tabular}
\end{table}

\begin{table}[t]
  \caption{ 
Ablation studies our approach for different values of `K' \label{Tab:Abl}}
\centering
\begin{tabular}{|l|l|l|l|l|}

\hline
\textit{Method} & \textit{mIoU} & \textit{Precision} & \textit{Recall} & \textit{F1} \\ \hline
\textbf{Ours K=100} & 0.25 & 0.50 & 0.35 & 0.34\\ \hline
\textbf{Ours K=50} & 0.32 & 0.48 & 0.51 & 0.42\\ \hline
\textbf{Ours K=30} & 0.33 & 0.49 & 0.49 & 0.43\\ \hline

\end{tabular}
\end{table}

\subsection{Experimental Results}

\begin{figure}[t!]
\centering
  \includegraphics[width=.4\textwidth]{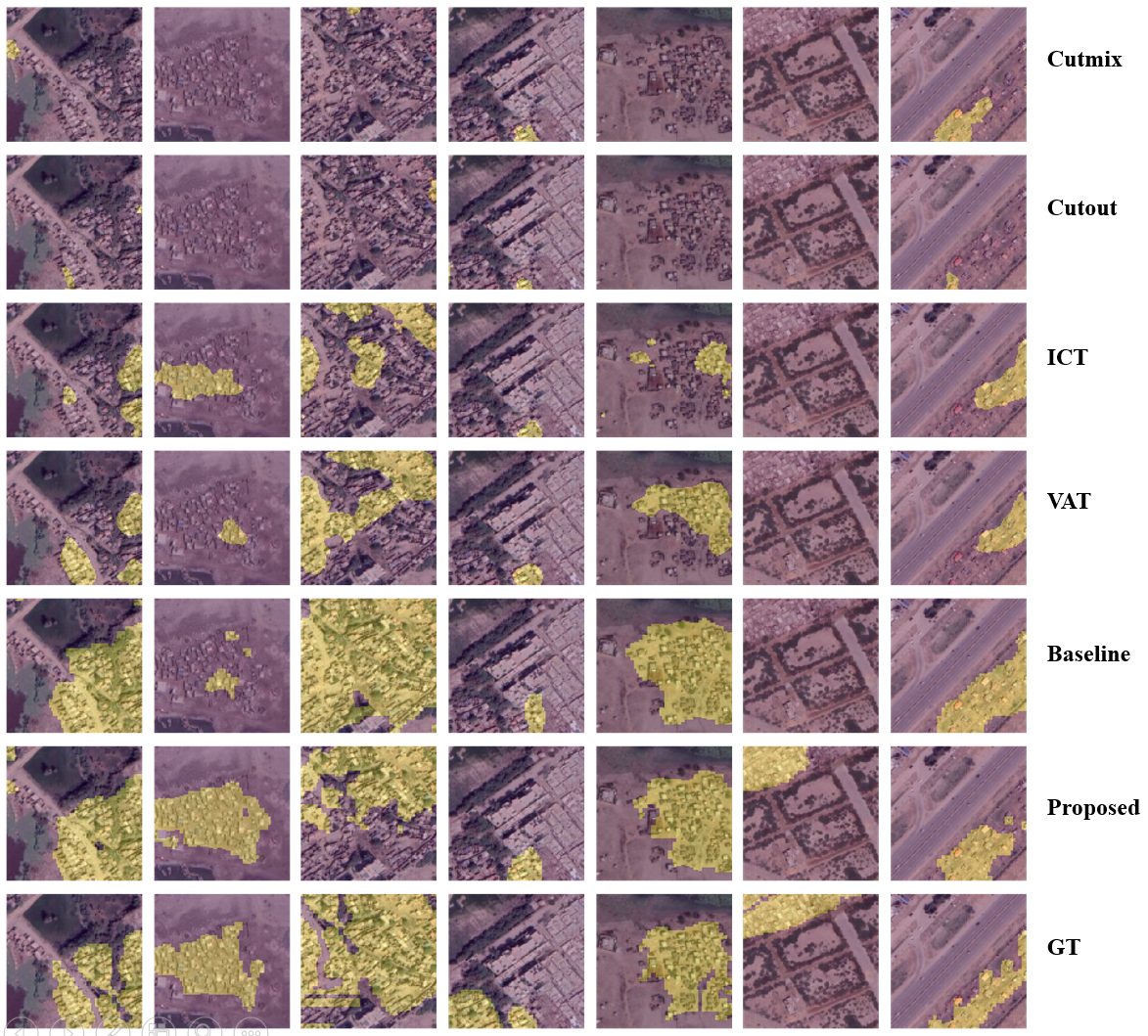}
  \caption{A comparison of the segmentation results of all the methods mentioned in Table \ref{Tab:Tcr}. The segmentation masks generated by the respective methods are overlayed over the satellite imagery.}\label{figure6}
\end{figure}

We have compared the proposed approach with several competitive baselines settings proposed in \cite{cutmix}. The authors in \cite{cutmix} used different regularization techniques such as  Cutout \cite{cutout}, Cutmix \cite{cutmix}, Interpolation Consistency Training (ICT) \cite{ict}, Virtual Adversarial Training (VAT) \cite{VAT}. For all the comparisons, we followed the experimental setup of \cite{cutmix} and used the same training, testing, and validation data as used in our experiment. The quantitative comparison of our approach in Table \ref{Tab:Tcr} demonstrates that the proposed approach significantly outperforms the baselines (Fig. \ref{figure6}).  {\textbf{Evaluation metrics: mIoU:} IOU is the ratio of the area of intersection and area of Union between prediction and ground truth. The mean value of IOU over all test images is denoted as mIOU. \textbf{Precision} is the ratio of true positives and all positives (true and false Positive). \textbf{Recall:} is the ratio of true positives and true positives plus false negatives. Finally, \textbf{F1-Score} combines the Precision and recall of a classifier into a single metric by taking their harmonic mean.}

\noindent\textbf{Ablation Studies:}
We evaluate our approach for different values of $K$ and demonstrate the results in Fig.~\ref{stopping} and  Table \ref{Tab:Abl}. The results demonstrate that the proposed approach outperforms the baselines for different values of $K$. Fig.~\ref{stopping} shows that for ${K}$ set to 30, our algorithm takes most iterations to reach the required threshold. However, this also results in the best results (mIoU), as indicated in Table~\ref{Tab:Tcr}. 
 
\noindent\textbf{Discussion:}
{While our proposed method performs significantly better against the other methods on segmentation metrics, the overall mIOU of is low below ideal. We consider the main contribution of our work is a methodology that automatically discovers temporary slum locations starting from minimal seed data. In our methodology the segmentation masks are just intermediate outputs, the problem we want to tackle in this work is `where to look for temporary slums?'. The only alternative to our work is to scan the whole city's satellite imagery manually. 
In the future, our approach can be improved in several ways. For instance, seed Selection can be improved if satellite imagery is available at more than two timestamps. Our methodology can be used in different geographical settings to identify the scale of temporary slums globally.}

\begin{figure}
\centering
  \includegraphics[width=0.4\textwidth]{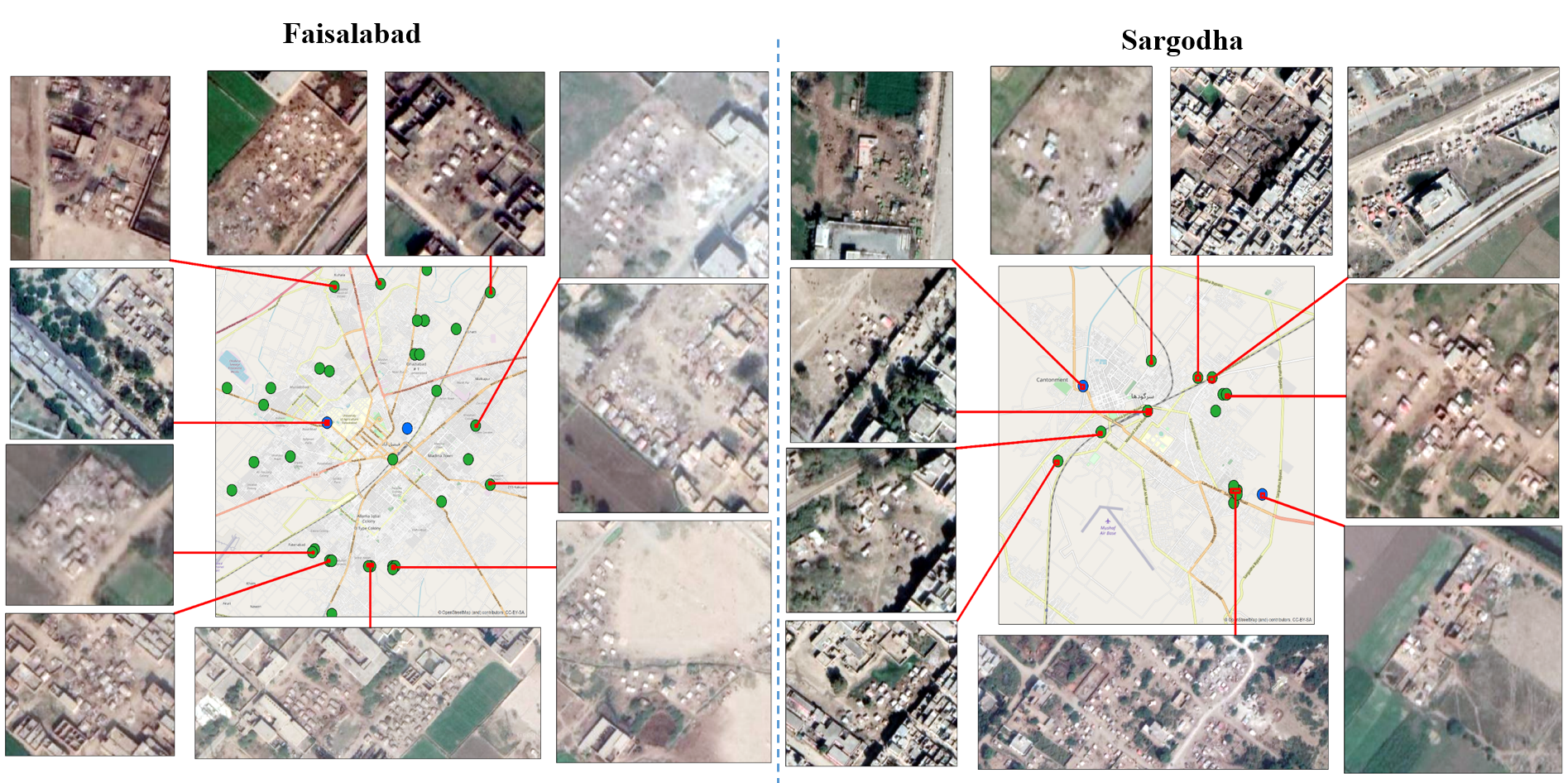}
  \caption{This figure shows the identified  {temporary} slum locations in Faisalabad and Sargodha, two of the 12 metropolitan cities of Pakistan that we covered. The green dots represent the visually verified locations, and the blue dots represent the locations where we are uncertain about our predictions. }\label{city}
\end{figure}

\section{Conclusion}

Finding temporary slums is a challenging task, as these areas exist mainly on the fringes of the cities, and the usual surveying methods would require significant resources both in terms of time and money.In our work, we have provided a semi-supervised learning-based solution that can be initiated from a small seed dataset. We used the temporal consistency of built-up structures to find those small seed images. We have generated temporary slums dataset for 12 metropolitan cities in Pakistan that can be updated and extended to other regions very easily. Our extensive experimental results and discussion validate the proposed
ideas and framework. We intend to use our works as a methodology that answers the question of "where to look for temporary slums?" and enables us to identify and visualize the scale of temporary slums.

\bibliographystyle{IEEEtran}
\bibliography{mybibfile}

\begin{thebibliography}{10}
\providecommand{\url}[1]{#1}
\csname url@samestyle\endcsname
\providecommand{\newblock}{\relax}
\providecommand{\bibinfo}[2]{#2}
\providecommand{\BIBentrySTDinterwordspacing}{\spaceskip=0pt\relax}
\providecommand{\BIBentryALTinterwordstretchfactor}{4}
\providecommand{\BIBentryALTinterwordspacing}{\spaceskip=\fontdimen2\font plus
\BIBentryALTinterwordstretchfactor\fontdimen3\font minus
  \fontdimen4\font\relax}
\providecommand{\BIBforeignlanguage}[2]{{%
\expandafter\ifx\csname l@#1\endcsname\relax
\typeout{** WARNING: IEEEtran.bst: No hyphenation pattern has been}%
\typeout{** loaded for the language `#1'. Using the pattern for}%
\typeout{** the default language instead.}%
\else
\language=\csname l@#1\endcsname
\fi
#2}}
\providecommand{\BIBdecl}{\relax}
\BIBdecl

\bibitem{slumAlmanac}
P.~T. Nairobi, ``Slum almanac 2015-2016: Tracking improvement in the lives of
  slum dwellers,'' \emph{Nairobi: UN Habitat. https://unhabitat.
  org/slum-almanac-2015-2016}, 2016.

\bibitem{un2015}
D.~UN, ``World urbanization prospects: The 2014 revision,'' \emph{United
  Nations Department of Economics and Social Affairs, Population Division: New
  York, NY, USA}, vol.~41, 2015.

\bibitem{slumanual1}
M.~Wurm and H.~Taubenb{\"o}ck, ``Detecting social groups from space--assessment
  of remote sensing-based mapped morphological slums using income data,''
  \emph{Remote Sensing Letters}, vol.~9, no.~1, pp. 41--50, 2018.

\bibitem{slumauto1}
M.~Kuffer, J.~Barros, and R.~V. Sliuzas, ``The development of a morphological
  unplanned settlement index using very-high-resolution (vhr) imagery,''
  \emph{Computers, Environment and Urban Systems}, vol.~48, pp. 138--152, 2014.

\bibitem{slumauto2}
I.~Baud, M.~Kuffer, K.~Pfeffer, R.~Sliuzas, and S.~Karuppannan, ``Understanding
  heterogeneity in metropolitan india: The added value of remote sensing data
  for analyzing sub-standard residential areas,'' \emph{International Journal
  of Applied Earth Observation and Geoinformation}, vol.~12, no.~5, pp.
  359--374, 2010.

\bibitem{slumGLCM1}
M.~Kuffer, K.~Pfeffer, R.~Sliuzas, and I.~Baud, ``Extraction of slum areas from
  vhr imagery using glcm variance,'' \emph{IEEE Journal of selected topics in
  applied earth observations and remote sensing}, vol.~9, no.~5, pp.
  1830--1840, 2016.

\bibitem{slumSVM1}
X.~Huang, H.~Liu, and L.~Zhang, ``Spatiotemporal detection and analysis of
  urban villages in mega city regions of china using high-resolution remotely
  sensed imagery,'' \emph{IEEE Transactions on Geoscience and Remote Sensing},
  vol.~53, no.~7, pp. 3639--3657, 2015.

\bibitem{slumDeep}
C.~Persello and A.~Stein, ``Deep fully convolutional networks for the detection
  of informal settlements in vhr images,'' \emph{IEEE geoscience and remote
  sensing letters}, vol.~14, no.~12, pp. 2325--2329, 2017.

\bibitem{wurm2019semantic}
M.~Wurm, T.~Stark, X.~X. Zhu, M.~Weigand, and H.~Taubenb{\"o}ck, ``Semantic
  segmentation of slums in satellite images using transfer learning on fully
  convolutional neural networks,'' \emph{ISPRS Journal of Photogrammetry and
  Remote sensing}, vol. 150, pp. 59--69, 2019.

\bibitem{simpleslum}
C.~Persello and A.~Stein, ``Deep fully convolutional networks for the detection
  of informal settlements in vhr images,'' \emph{IEEE geoscience and remote
  sensing letters}, vol.~14, no.~12, pp. 2325--2329, 2017.

\bibitem{imbSlums}
T.~Stark, M.~Wurm, H.~Taubenb{\"o}ck, and X.~Zhu, ``Slum mapping in imbalanced
  remote sensing datasets using transfer learned deep features,'' \emph{2019
  Joint Urban Remote Sensing Event (JURSE)}, 2019.

\bibitem{desa2016transforming}
U.~Desa \emph{et~al.}, ``Transforming our world: The 2030 agenda for
  sustainable development,'' 2016.

\bibitem{mustafainformal}
M.~Mustafa, ``The informal settlements of lahore: Understanding the role of
  informal katchi abadi in the context of affordable housing,''
  \emph{Department of Architecture \& Planning, NED University of Engineering
  \& Technology, City Campus Maulana Din Muhammad Wafai Road, Karachi.}, p.~38.

\bibitem{VF}
K.~Murtaza, S.~Khan, and N.~M. Rajpoot, ``Villagefinder: Segmentation of
  nucleated villages in satellite imagery.'' in \emph{BMVC}.\hskip 1em plus
  0.5em minus 0.4em\relax Citeseer, 2009, pp. 1--11.

\bibitem{unet}
O.~Ronneberger, P.~Fischer, and T.~Brox, ``U-net: Convolutional networks for
  biomedical image segmentation,'' in \emph{International Conference on Medical
  image computing and computer-assisted intervention}.\hskip 1em plus 0.5em
  minus 0.4em\relax Springer, 2015, pp. 234--241.

\bibitem{resnet}
K.~He, X.~Zhang, S.~Ren, and J.~Sun, ``Deep residual learning for image
  recognition,'' in \emph{Proceedings of the IEEE conference on computer vision
  and pattern recognition}, 2016, pp. 770--778.

\bibitem{imagenet}
J.~Deng, W.~Dong, R.~Socher, L.-J. Li, K.~Li, and L.~Fei-Fei, ``Imagenet: A
  large-scale hierarchical image database,'' in \emph{2009 IEEE conference on
  computer vision and pattern recognition}.\hskip 1em plus 0.5em minus
  0.4em\relax Ieee, 2009, pp. 248--255.

\bibitem{cutmix}
G.~French, S.~Laine, T.~Aila, M.~Mackiewicz, and G.~Finlayson,
  ``Semi-supervised semantic segmentation needs strong, varied perturbations,''
  in \emph{British Machine Vision Conference}, no.~31, 2020.

\bibitem{ict}
V.~Verma, K.~Kawaguchi, A.~Lamb, J.~Kannala, Y.~Bengio, and D.~Lopez-Paz,
  ``Interpolation consistency training for semi-supervised learning,''
  \emph{arXiv preprint arXiv:1903.03825}, 2019.

\bibitem{VAT}
T.~Miyato, S.-i. Maeda, M.~Koyama, and S.~Ishii, ``Virtual adversarial
  training: a regularization method for supervised and semi-supervised
  learning,'' \emph{IEEE transactions on pattern analysis and machine
  intelligence}, vol.~41, no.~8, pp. 1979--1993, 2018.

\bibitem{cutout}
T.~DeVries and G.~W. Taylor, ``Improved regularization of convolutional neural
  networks with cutout,'' \emph{arXiv preprint arXiv:1708.04552}, 2017.

\end{thebibliography}

\end{document}